\documentclass{article}

\usepackage{PRIMEarxiv}

\usepackage[utf8]{inputenc}
\usepackage[T1]{fontenc}
\usepackage{lmodern}
\usepackage[hidelinks]{hyperref}
\usepackage{url}
\usepackage{booktabs}
\usepackage{amsmath, amssymb}
\usepackage{siunitx}
\usepackage{nicefrac}
\usepackage{microtype}
\usepackage{graphicx}
\usepackage{xcolor}
\usepackage{enumitem}
\usepackage{rotating}

\usepackage{float}
\usepackage{placeins}

\usepackage{booktabs}   % \toprule, \midrule, \bottomrule
\usepackage{multirow}   % \multirow in Table 1
\usepackage{longtable}  % required because your file uses longtable

\usepackage{textcomp}       % common text symbols
\usepackage{newunicodechar} % map specific Unicode to TeX macros

\providecommand{\DeclareUnicodeCharacter}[2]{}
\DeclareUnicodeCharacter{2192}{\textrightarrow} % →  (use $\to$)
\DeclareUnicodeCharacter{2013}{--}              % –  (en dash)
\DeclareUnicodeCharacter{2014}{---}             % —  (em dash, just in case)
\DeclareUnicodeCharacter{00D7}{\(\times\)}      % ×
\DeclareUnicodeCharacter{00B1}{\(\pm\)}         % ±
\DeclareUnicodeCharacter{2026}{\ldots}          % …
\DeclareUnicodeCharacter{2212}{-}               % − (unicode minus)

\newunicodechar{→}{\textrightarrow}
\newunicodechar{–}{--}          % en dash
\newunicodechar{—}{---}         % em dash (if any)
\newunicodechar{×}{\(\times\)}  % or {\texttimes}
\newunicodechar{±}{\(\pm\)}
\newunicodechar{…}{\ldots}
\newunicodechar{−}{-}           % Unicode minus → ASCII hyphen (or {$-$})

\usepackage{array}      % new column types
\usepackage{ltablex}    % longtable + tabularx
\keepXColumns           % keep X column settings across head/foot
\usepackage{booktabs}
\usepackage{multirow}
\usepackage{longtable}  % loaded by ltablex, harmless to keep
\graphicspath{{media/}}

\title{Anchors in the Machine: Behavioral and Attributional Evidence of Anchoring Bias in LLMs
}

\author{
  Felipe Valencia-Clavijo \\
  Dataplicada \\
  \texttt{feval@dataplicada.com} \\
}

\begin{document}
\maketitle
\vspace{-2.5em} % adjust value until it looks right

\begin{abstract}
Large language models (LLMs) are increasingly studied as behavioral subjects (e.g., synthetic participants or users) and decision systems (e.g., LLM-as-a-judge), yet it remains unclear how observed cognitive biases relate to changes in model output probabilities. \emph{Anchoring bias}, a classic human judgment bias, provides a useful test case. Prior research largely relies on generated responses, leaving distributional effects and input-field contributions less explored. This paper advances the study of anchoring in LLMs through three contributions: (1) a log-probability-based behavioral analysis examining how numerical anchors shift distributions over candidate outputs, using alternative questions and anchor pairs alongside a positive control to reduce reliance on potentially memorized stimuli; (2) exact Banzhaf-value attribution over structured prompt fields to quantify anchor contributions to log-probabilities; and (3) an Anchoring Bias Sensitivity Score (ABSS) integrating behavioral and attributional evidence across six open-source models. Results reveal predominantly positive behavioral anchoring effects, varying substantially across models. Gemma-2B, Phi-2, and Llama-2-7B achieve the highest ABSS rankings, whereas other models exhibit weaker behavioral--attributional alignment despite sometimes showing substantial behavioral shifts. Attributional effects also vary across prompts, indicating that behavioral anchoring does not necessarily imply stronger anchor-field contributions. These findings raise questions about LLM reliability as synthetic participants and potential vulnerabilities to numerical cues in automated evaluation and decision-making. More broadly, the framework connects behavioral science with explainable AI, particularly interpretability, and AI safety, providing a reproducible approach to investigating cognitive-bias-like effects in LLMs.
\end{abstract}

% Keywords

% Optional notation/macros
\newcommand{\softEV}{\mathrm{SoftEV}}
\newcommand{\LLM}{\mathrm{LLM}}
\newcommand{\shap}{\phi}
\newcommand{\deltasoft}{\Delta\!\softEV}

% =========================
% ---------- Helpers to reuse the same footnote multiple times ----------
\newcommand{\footnoteremember}[2]{\footnote{#2\label{#1}}}
\newcommand{\footnoterecall}[1]{\textsuperscript{\ref{#1}}}
% Reset footnote counter so Methods starts at 1 (title \thanks uses footnote 1)
\setcounter{footnote}{0}
% ----------------------------------------------------------------------

% =========================
\raggedbottom
\enlargethispage{2\baselineskip}

\begin{figure}[!htbp]
    \centering
    \includegraphics[width=0.86\textwidth,keepaspectratio]{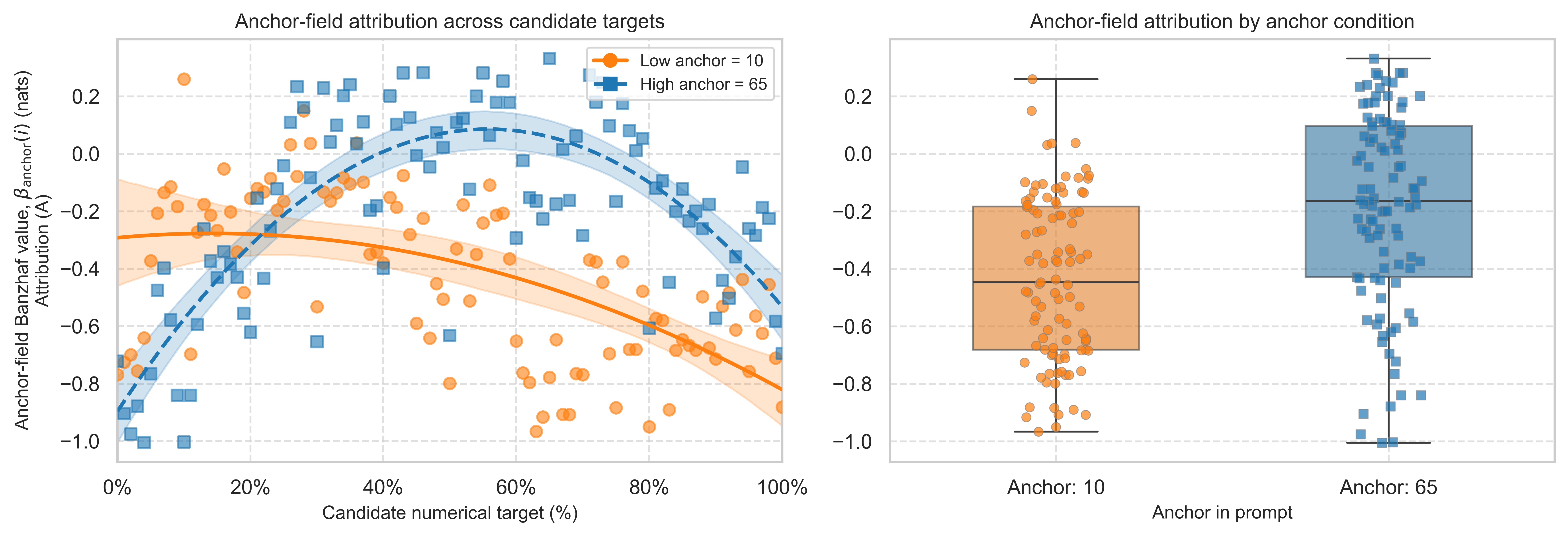}
    \caption{Replication of Tversky and Kahneman's anchoring experiment~\cite{doi:10.1126/science.185.4157.1124} in GPT-2 using exact Banzhaf-value attribution.}
    \label{fig:replicate_1974}
\end{figure}

% =========================
\section{Introduction}

Two complementary research streams have emerged at the intersection of decision sciences and artificial intelligence. One treats large language models (LLMs) as \emph{subjects}, examining whether they reproduce human cognitive phenomena and whether they might serve as synthetic participants or users in behavioral research and social simulation \cite{cui2025large,mooney2025llmagentsbehaviorallycoherent}. The other draws on neuroscience, behavioral economics, and psychology to investigate LLMs as decision systems in their own right, including applications such as \emph{LLM-as-a-judge} and automated evaluation, with the aim of identifying failure modes and improving reliability and safety \cite{binz2023cognitive,Ahmed_Jaźwińska_Ahlawat_Winecoff_Wang_2024}. Both perspectives are increasingly relevant as LLMs are integrated into consequential evaluation and decision-making processes through autonomous and semi-autonomous agents.

A persistent challenge for both streams is the limited transparency of LLM predictions. Research in interpretable and explainable AI has developed approaches ranging from behavioral input-output analysis and feature attribution to concept-based probing and mechanistic investigations of internal computations \cite{molnar2025,saphra2024mechanistic,bereska2024mechanisticinterpretabilityaisafety}. Although these approaches operate at different levels, they share the objective of improving our understanding of model behavior. However, a gap remains between studies that identify behavioral patterns in generated responses and those that quantify how particular input components influence model predictions.

Cognitive biases provide a useful framework for investigating this gap. Decades of behavioral research have documented systematic deviations from normative judgment and decision-making \cite{TVERSKY1973207,doi:10.1126/science.185.4157.1124,tversky1981framing}. Recent studies suggest that LLMs reproduce several psychology-inspired effects \cite{cui2025large,binz2023cognitive}, while related research has identified systematic sensitivities and failure modes in model predictions \cite{jones2022capturingfailureslargelanguage,liu2021makesgoodincontextexamples,zhao2021calibrateuseimprovingfewshot}. For LLMs-as-subjects research, these findings raise questions about the extent to which apparent human-like biases generalize across experimental conditions and whether synthetic participants can reliably approximate human judgments. For LLMs-as-decision-systems research, including LLM-as-a-judge applications, they highlight potential vulnerabilities to contextual cues that may affect the reliability of automated evaluations and model-assisted decisions.

This paper focuses on \emph{anchoring bias}, the tendency for numerical judgments to be influenced by an initial reference value, even when that value is arbitrary or uninformative. In the classic experiment of Tversky and Kahneman~\cite{doi:10.1126/science.185.4157.1124}, participants estimating the percentage of African countries in the United Nations reported higher estimates after exposure to a high anchor (65) than after a low anchor (10). Subsequent research has documented anchoring-like effects in LLMs across different models and tasks \cite{suri2023largelanguagemodelsdecision,lou2024anchoringbiaslargelanguage,stureborg2024largelanguagemodelsinconsistent}. However, much of this evidence relies on generated responses, leaving less explored how numerical anchors change the distribution of probabilities over candidate answers and how the anchor field contributes to individual output scores.

I address these questions by combining behavioral distributional analysis with exact game-theoretic attribution. On the behavioral side, I evaluate sequence log-probabilities over a fixed set of numerical answers and normalize them into a categorical distribution. This allows anchoring effects to be examined through changes in the full evaluated output distribution rather than only through sampled responses. On the attributional side, I use exact Banzhaf values over four structured prompt fields to quantify the anchor field's average marginal contribution to candidate-answer log-probabilities. Unlike classical Shapley values, Banzhaf attribution~\cite{banzhaf1965weighted} assigns equal weight to all coalitions excluding the evaluated field. The four-field representation permits exhaustive evaluation of all 16 field subsets, avoiding Monte Carlo approximation of the attribution values. This approach builds on game-theoretic interpretability research \cite{shapley1953value,banzhaf1965weighted,mohammadi2024explaininglargelanguagemodels,goldshmidt2024tokenshapinterpretinglargelanguage,NIPS2017_7062}, while adopting a distinct attribution measure suited to the present experimental design.

Empirically, I evaluate six LLMs (GPT-2, GPT-Neo-125M, Falcon-RW-1B, Gemma-2B, Phi-2, and Llama-2-7B) across standard and alternative anchor pairs, using controlled prompt structures. I additionally introduce an Anchoring Bias Sensitivity Score (ABSS) that integrates behavioral shifts, Banzhaf attributional changes, statistical evidence, robustness diagnostics, and directional concordance. The results reveal heterogeneous anchoring sensitivity across models: positive behavioral shifts are common, but their magnitudes vary, and attributional responses do not always align with behavioral direction. These findings show that anchoring-like behavior in LLMs can be accompanied by measurable changes in both output probability distributions and anchor-field attribution, without implying that the models reproduce the cognitive mechanisms underlying human anchoring. Together, the behavioral and attributional analyses provide complementary evidence for characterizing context sensitivity in LLM predictions.

% =========================
\section{Related Work}

\subsection{Interpretable AI Paradigms}

Bereska and Gavves~\cite{bereska2024mechanisticinterpretabilityaisafety} outline four main paradigms for interpreting model behavior:

\begin{enumerate}
    \item \textbf{Behavioral interpretability} --- treating models as black boxes and examining input-output relationships.
    \item \textbf{Attributional interpretability} --- quantifying the influence of input features on model predictions.
    \item \textbf{Concept-based interpretability} --- examining internal representations associated with higher-level concepts.
    \item \textbf{Mechanistic interpretability} --- investigating the internal components and computational mechanisms underlying model behavior.
\end{enumerate}

This study combines the first two paradigms. Behavioral analysis examines how numerical anchors shift probability distributions over candidate outputs, while attributional analysis uses exact Banzhaf values to quantify the marginal contribution of the anchor field to model log-probabilities. This combination extends beyond surface-level output comparisons without making claims about internal representations or computational mechanisms.

\subsection{Cognitive Biases}

Cognitive biases are systematic patterns of deviation from normative models of judgment and decision-making, prominently investigated in the pioneering work of Tversky and Kahneman. Their early study of the availability heuristic showed that judgments of frequency and probability can be influenced by how readily examples come to mind~\cite{TVERSKY1973207}. Their subsequent work identified representativeness, availability, and anchoring as central heuristics associated with systematic judgment errors~\cite{doi:10.1126/science.185.4157.1124}. Later, they demonstrated how different presentations of equivalent decision problems can produce systematic shifts in preferences, establishing the importance of framing effects~\cite{tversky1981framing}. These findings contributed to the foundations of behavioral economics by demonstrating predictable departures from normative decision-making. More recently, they have motivated research examining whether LLMs reproduce similar patterns of judgment and whether such patterns reflect human-like responses or distinct properties of model behavior.

\subsection{Cognitive Bias in LLMs}

Cognitive biases in LLMs are increasingly examined through psychology-inspired experiments. Cui et al.~\cite{cui2025large} report that models reproduce several established cognitive biases, although their magnitudes and statistical properties may differ from those observed in humans. Similarly, Binz and Schulz~\cite{binz2023cognitive} found that GPT-3 exhibited human-like patterns in several cognitive tasks, alongside limitations in reasoning and sensitivity to experimental conditions. Together, these findings suggest that behavioral similarities between humans and LLMs should not be interpreted as evidence of equivalent underlying processes. Moreover, behavioral replications alone do not quantify how individual prompt components contribute to the probabilities assigned to candidate outputs.

\subsection{Anchoring Bias in LLMs}

Among the cognitive biases examined in LLMs, anchoring has received particular attention. Experimental studies indicate that numerical cues can systematically influence model-generated judgments, even when those cues are not directly informative about the target quantity.

Suri et al.~\cite{suri2023largelanguagemodelsdecision} investigate anchoring through comparisons involving ChatGPT-3.5, GPT-4, and human participants, documenting shifts in estimates between high- and low-anchor conditions. Similarly, Lou and Sun~\cite{lou2024anchoringbiaslargelanguage} examine anchoring across GPT-3.5, GPT-4, and GPT-4o, highlighting differences in susceptibility across models and the limitations of prompt-based mitigation strategies. Stureborg et al.~\cite{stureborg2024largelanguagemodelsinconsistent} extend this research to multi-attribute evaluation tasks, showing that earlier assessments can influence subsequent ratings. Such sequential dependencies raise concerns about the reliability of LLMs when used as evaluators.

These studies establish anchoring as a relevant behavioral phenomenon in LLMs. However, much of the evidence relies on generated responses or reported judgments, leaving opportunities to investigate how anchoring changes the probability distribution over possible outputs and how the anchor field contributes to those probabilities.

\subsection{Other Related Cognitive Biases in LLMs}

Related experimental work has documented biases in LLM evaluation and decision-making. Wang et al.~\cite{wang2023largelanguagemodelsfair} identify positional biases in LLM-based evaluators, demonstrating sensitivity to the ordering of candidate responses. Chen et al.~\cite{chen2024aicognitivelybiasedexploratory} examine threshold priming in information retrieval assessments, where preceding relevance scores influence subsequent evaluations. Sumita et al.~\cite{sumita2024cognitivebiaseslargelanguage} investigate multiple cognitive biases, including order-related effects.

Moving beyond behavioral evaluation, Li and Gao~\cite{li2025anchoredanswersunravellingpositional} apply mechanistic interpretability to multiple-choice answering in GPT-2, investigating an internal preference for the first response option. These studies illustrate that context, presentation, and ordering can influence LLM predictions. They also highlight the value of distinguishing observable behavioral effects, input attribution, and internal mechanistic explanations.

\subsection{Game-Theoretic Attribution in LLMs}

Behavioral experiments establish whether model predictions change under different inputs, but do not by themselves quantify the contribution of individual prompt components. Cooperative-game-theoretic attribution methods provide a complementary approach by treating input components as players and model predictions as payoffs.

The Shapley value, introduced by Shapley~\cite{shapley1953value}, allocates a payoff among players by averaging their marginal contributions using coalition-size-dependent weights. In machine learning, this principle underlies several approaches to feature attribution, including SHAP. Mohammadi~\cite{mohammadi2024explaininglargelanguagemodels} develops a Shapley-based framework for interpreting LLM decisions by attributing model outputs to prompt components. The study also identifies ``token noise,'' where seemingly uninformative prompt elements disproportionately influence model decisions. Horovicz and Goldshmidt~\cite{goldshmidt2024tokenshapinterpretinglargelanguage} introduce \textit{TokenSHAP}, which uses Monte Carlo Shapley-value estimation to quantify the importance of prompt tokens or substrings. The SHAP framework also provides approaches for attributing text-model outputs to input features, illustrating the broader applicability of cooperative game theory to language-model interpretability~\cite{NIPS2017_7062}.

An alternative to classical Shapley attribution is the Banzhaf value~\cite{banzhaf1965weighted}. Both approaches quantify marginal contributions across coalitions, but they differ in how these contributions are aggregated. Whereas Shapley values weight coalitions according to their size, Banzhaf values assign equal weight to every coalition excluding the evaluated feature. Consequently, Banzhaf attribution measures the average marginal contribution under uniformly weighted field-presence combinations and does not generally satisfy the Shapley efficiency property.

This distinction motivates the attributional approach adopted in the present study. Rather than estimating token-level Shapley contributions, I compute exact Banzhaf values over four structured prompt fields, using the log-probability of each candidate numerical output as the payoff. The small feature space permits exhaustive evaluation of all $2^4=16$ field subsets without Monte Carlo approximation.

Combining these attributions with behavioral analysis of normalized output distributions allows anchoring to be examined from two complementary perspectives: changes in expected estimates and changes in the anchor field's marginal contribution to model log-probabilities. This provides a field-level attributional account of anchoring sensitivity without claiming to identify the model's internal computational mechanisms.

% =========================
\section{Methods}

I study anchoring in large language models using a controlled experimental design. Following the interpretability paradigms discussed by Bereska and Gavves~\cite{bereska2024mechanisticinterpretabilityaisafety}, I examine two complementary forms of evidence: (i) behavioral (B) shifts in the probability distribution over numerical targets when a low anchor is replaced with a high anchor, and (ii) attributional (A) changes in the contribution of the \emph{anchor} field to the model's log-probabilities for those same targets. The attributional analysis uses exact Banzhaf values over structured prompt fields, providing a game-theoretic measure of marginal input contribution. I do not attempt concept-based (C) or mechanistic (M) interpretability; instead, I examine how anchor manipulations relate to observable distributional shifts and attributional changes.

\subsection{Design, stimuli, and hypotheses}

\paragraph{Positive-control replication (V0; Tversky \& Kahneman).}
As a positive control, I replicate the classic ``African countries in the UN'' anchoring experiment reported by Tversky and Kahneman~\cite{doi:10.1126/science.185.4157.1124}. Inspired by the prompt-template approach of Mohammadi~\cite{mohammadi2024explaininglargelanguagemodels}, I use a three-sentence prompt containing a single numerical anchor. V0 is retained as a positive control but excluded from aggregate analyses and rankings to reduce potential contamination from a canonical experiment that may have appeared in pre-training data. All other stimuli are listed in Appendix~\ref{app:prompts}.

\paragraph{Template (fixed structure).}
Prompts are rendered using a Jinja template with four structured fields \texttt{\{scene, comparative, absolute, anchor\}}:

\begin{verbatim}
{{ scene }}{{ anchor }}.

{{ comparative }}{{ anchor }}?

{{ absolute }}
\end{verbatim}

\paragraph{V0 example (rendered).}
\begin{itemize}[leftmargin=*]
    \item \textbf{Low anchor (10):}
    \begin{quote}
        \emph{The roulette wheel landed on 10.}\\
        \emph{Is the percentage of African countries in the United Nations larger or smaller than 10?}\\
        \emph{What is your best guess of the percentage of African countries in the UN?}
    \end{quote}
    \item \textbf{High anchor (65):}
    \begin{quote}
        \emph{The roulette wheel landed on 65.}\\
        \emph{Is the percentage of African countries in the United Nations larger or smaller than 65?}\\
        \emph{What is your best guess of the percentage of African countries in the UN?}
    \end{quote}
\end{itemize}

\paragraph{Question families and anchor regimes.}
Beyond V0, I include different questions while preserving the same prompt structure to examine anchoring across content (e.g., Asian, South American, English-speaking, EU, and French-speaking countries in the UN).

I consider two regimes: (i) a \emph{standard-anchors} regime (S), which reuses the pair (10, 65) across five questions; and (ii) a \emph{different-anchors} regime (D), which varies the numerical pair while preserving the same 55-point separation (15--70, 20--75, $\ldots$, 35--90).

Maintaining a constant numerical separation facilitates comparison of behavioral effect magnitudes across anchor pairs, while changing the absolute values reduces reliance on a single potentially familiar numerical pair. The full stimuli are provided in Appendix~\ref{app:prompts}.

\paragraph{Hypotheses and direction calls.}
\begin{itemize}[leftmargin=*]
    \item \textbf{Behavior (B):}
    \[
    H_{0}^{B}:\; \mathbb{P}(Y\mid\text{high})=\mathbb{P}(Y\mid\text{low}), \qquad
    H_{1}^{B}:\; \mathbb{P}(Y\mid\text{high})\neq\mathbb{P}(Y\mid\text{low}).
    \]
    I interpret \textbf{B+} as a behavioral shift aligned with anchoring bias, where the soft expected value increases under the high anchor; \textbf{B--} indicates a shift in the opposite direction, and \textbf{B0} indicates no directional difference. Behavioral direction is determined from the difference in soft expected values, while statistical evidence is assessed using paired target-level log-probability differences. Significance is denoted by $*$ ($p<0.10$), $**$ ($p<0.05$), and $***$ ($p<0.01$).

    \item \textbf{Attribution (A):}
    \[
    H_{0}^{A}:\; \mathbb{E}[\beta_{\mathrm{anchor}}\mid\text{high}]=\mathbb{E}[\beta_{\mathrm{anchor}}\mid\text{low}],
    \]
    \[
    H_{1}^{A}:\; \mathbb{E}[\beta_{\mathrm{anchor}}\mid\text{high}]\neq\mathbb{E}[\beta_{\mathrm{anchor}}\mid\text{low}].
    \]
    Here, $\beta_{\mathrm{anchor}}$ denotes the Banzhaf attribution of the anchor field. I interpret \textbf{A+} as a positive difference in mean anchor-field Banzhaf attribution between high- and low-anchor conditions, indicating a larger marginal contribution under the high anchor. \textbf{A--} indicates the reverse, while \textbf{A0} denotes no directional difference. Attribution differences are expressed in log-probability units (nats). Exponentiating a difference provides a multiplicative interpretation of the associated log-probability contribution; for example, a difference of approximately $0.69$ nats corresponds to a multiplier of approximately $\times 2$. Attributional significance follows the same $*$, $**$, and $***$ thresholds as the behavioral analysis.
\end{itemize}

\subsection{End-to-end procedure}
\label{sec:procedure}

\begin{enumerate}[leftmargin=*]

    \item \label{step:setup}\textbf{Setup.} I load each model and tokenizer using Hugging Face, fix random seeds, and evaluate target sequences in teacher-forced mode without sampling.\footnoteremember{fn:decoding}{\textbf{Decoding.} I score \texttt{prompt + " " + target} by summing token-level log-probabilities for the target string. Causal alignment uses logits at position $t-1$ to score token $t$.}

    \item \label{step:render}\textbf{Render prompts.} For each variation $v$ and anchor condition $a\in\{\text{low},\text{high}\}$, I render the three-sentence prompt using the fixed template. Appendix~\ref{app:prompts} lists all non-control variations.

    \item \label{step:score}\textbf{Score all targets.} I restrict the evaluated outputs to the fixed set $\{0,1,\ldots,100\}$, represented as strings ``$i\%$''. For each target $y_i$, I compute its sequence log-probability:
    \[
    \ell_i=\log P(y_i\mid\text{prompt}),
    \]
    summing the log-probabilities across the target's constituent tokens.\footnoterecall{fn:decoding}

    \item \label{step:norm}\textbf{Normalize to a categorical distribution.} From the log-probabilities $\{\ell_i\}$, I construct a normalized distribution over the 101 candidate targets using log-sum-exp:
    \[
    p_i=\exp\left(\ell_i-\operatorname{logsumexp}_{j}\ell_j\right), \qquad \sum_i p_i=1.
    \]
    Numerical stability is maintained using log-sum-exp and appropriate floating-point casting where required by the model backend.

    \item \label{step:softev}\textbf{Behavioral summary (SoftEV).} I summarize the normalized distribution using its soft expected value:
    \[
    \mathrm{SoftEV}=\sum_{i=0}^{100} i\,p_i.
    \]
    The behavioral shift for each variation is
    \[
    \Delta\mathrm{EV}=\mathrm{SoftEV}_{\mathrm{high}}-\mathrm{SoftEV}_{\mathrm{low}}.
    \]
    I additionally estimate a 95\% predictive interval for the mean of $n=100$ draws from the implied categorical distribution, using $B=5000$ simulated resamples.\footnoteremember{fn:predictive}{\textbf{Predictive band.} The interval corresponds to the 2.5th--97.5th percentiles of the means of $n$ simulated draws from the normalized categorical distribution. It represents predictive variation under the model-implied distribution, not a confidence interval for a human population parameter.}

    \item \label{step:bt}\textbf{Primary behavioral test (B).} Using the log-probabilities from Step~\ref{step:score}, I compute the paired differences for each target:
    \[
    d_i=\ell_i^{(\mathrm{high})}-\ell_i^{(\mathrm{low})}.
    \]
    I apply a two-sided paired $t$-test to the target-level log-probabilities under the two anchor conditions. This test characterizes differences in the evaluated log-probability profiles. Because the target scores arise from the same model evaluations and are not independent experimental observations, the resulting $p$-values are interpreted as conditional diagnostics rather than independent-sample inference. The behavioral direction (\textbf{B+}, \textbf{B--}, or \textbf{B0}) is determined using $\Delta\mathrm{EV}$, while significance symbols are assigned using the log-probability test.

    \item \label{step:robust}\textbf{Robustness for B.} I supplement the paired $t$-test with two additional target-level diagnostics on the differences $\{d_i\}$: (i) a two-sided Wilcoxon signed-rank test using Pratt's treatment of zeros; and (ii) a sign-flip permutation test using the mean difference as its statistic.\footnoteremember{fn:wilcoxon}{\textbf{Wilcoxon.} Two-sided signed-rank test using Pratt handling of zero differences.}\footnoteremember{fn:perm}{\textbf{Permutation.} Random Rademacher sign-flips applied to the target-level differences, using the mean as the test statistic and $10{,}000$ permutations.} These tests assess the sensitivity of the target-level results to alternative test statistics, without eliminating the dependence inherent in the evaluated probability profiles.

    \item \label{step:attr}\textbf{Attribution (A): exact Banzhaf values.} I quantify the marginal contribution of the anchor field using exact Banzhaf attribution over the four structured prompt fields. Let
    \[
    N=\{\mathrm{scene},\mathrm{comparative},\mathrm{absolute},\mathrm{anchor}\}
    \]
    denote the set of fields, with $n=4$. For each variation $v$, anchor condition $a$, and target $y_i$, I define the log-probability payoff:
    \[
    v_i(S)=\log P\left(y_i\mid\text{prompt with fields }S\right),
    \]
    where $S\subseteq N$ denotes a subset of the available prompt fields. Absent fields are replaced with empty strings while preserving the prompt rendering procedure. The Banzhaf value for the anchor field is
    \[
    \boxed{
    \beta_{\mathrm{anchor}}(i)=\frac{1}{2^{n-1}}\sum_{S\subseteq N\setminus\{\mathrm{anchor}\}}\left[v_i(S\cup\{\mathrm{anchor}\})-v_i(S)\right]
    }
    \]
    Unlike classical Shapley values, which weight marginal contributions according to coalition size, Banzhaf values assign equal weight to all coalitions excluding the evaluated field~\cite{banzhaf1965weighted}. I evaluate all $2^4=16$ possible field subsets and compute the marginal contributions exactly. This avoids Monte Carlo approximation of the coalition-based attribution.

    For each anchor condition, I summarize the resulting target-level Banzhaf values using their arithmetic mean:
    \[
    \overline{\beta}_{a}=\frac{1}{K}\sum_{i=1}^{K}\beta_{\mathrm{anchor}}^{(a)}(i), \qquad K=101,
    \]
    where the mean is taken over the evaluated targets with valid attributions. The attributional shift is
    \[
    \Delta\beta=\overline{\beta}_{\mathrm{high}}-\overline{\beta}_{\mathrm{low}}.
    \]
    I express this shift using the attribution multiplier
    \[
    M_A=e^{\Delta\beta}.
    \]
    A multiplier above one indicates a positive attributional shift, whereas a multiplier below one indicates a negative shift. Attributional evidence is assessed using a two-sided paired $t$-test on target-level Banzhaf values under the high- and low-anchor conditions. As in the behavioral analysis, these tests are interpreted conditionally on the evaluated targets.

    Prior attribution approaches, including Mohammadi~\cite{mohammadi2024explaininglargelanguagemodels} and TokenSHAP~\cite{goldshmidt2024tokenshapinterpretinglargelanguage}, motivate the use of cooperative-game-theoretic measures for interpreting language models. My approach differs by using equal-weight Banzhaf attribution, a structured four-field representation, and log-probability payoffs, making full coalition enumeration tractable. The resulting attributions quantify field-level marginal contributions under the specified payoff and coalition procedure, rather than identifying internal computational mechanisms.

    As an illustrative example, Figure~\ref{fig:behavior_attribution_3d} (Appendix~\ref{app:behavior_attribution_example}) visualizes the relationship between normalized target-level log-probabilities and anchor-field Banzhaf values for \textit{gemma-2b} (V2-S), comparing the low (10) and high (65) anchor conditions. The figure provides a descriptive representation of behavioral and attributional variation across candidate outputs.

    \item \label{step:agg}\textbf{Aggregation across variations and models: Anchoring Bias Sensitivity Score (ABSS).} I construct a model-level Anchoring Bias Sensitivity Score (ABSS) combining behavioral and attributional evidence, statistical significance, robustness, and directional concordance. For each non-control variation, I define the signed behavioral component as
    \[
    S_B=\operatorname{sign}\left(\frac{\Delta\mathrm{EV}}{100}\right)\left|\frac{\Delta\mathrm{EV}}{100}\right|,
    \]
    and the signed attribution component as
    \[
    S_A=\operatorname{sign}(\Delta\beta)\tanh\left(|\Delta\beta|\right).
    \]
    Here, $\Delta\mathrm{EV}$ denotes the difference between high- and low-anchor SoftEV values, and $\Delta\beta$ denotes the difference in mean anchor-field Banzhaf attribution.

    I map the behavioral and attributional test $p$-values to weights:
    \[
    w(p)=\operatorname{clip}\left(\frac{-\log_{10}(p)}{3},0,1\right).
    \]
    Missing or invalid $p$-values receive zero weight in the implementation. I construct a robustness factor using the Wilcoxon and permutation test weights:
    \[
    \rho=0.5+0.5\,\operatorname{mean}\left[w(p_{\mathrm{Wil}}),w(p_{\mathrm{Perm}})\right].
    \]
    The concordance term $c\in\{-1,0,+1\}$ rewards agreement and penalizes disagreement between behavioral and attributional directions. Specifically, when both significance weights exceed $0.2$ and both effects have nonzero directions, $c=+1$ if their signs agree and $c=-1$ otherwise. In all other cases, $c=0$.

    With $\alpha=\beta=1$ and $\lambda_{\mathrm{conc}}=0.15$, the per-variation score is
    \[
    \boxed{
    \mathrm{ABSS}_{v}=\rho\left[\alpha S_B w(p_{\log})+\beta S_A w(p_{\mathrm{attr}})\right]+\lambda_{\mathrm{conc}}c
    }
    \]
    where $p_{\log}$ denotes the behavioral log-probability test $p$-value and $p_{\mathrm{attr}}$ denotes the attributional Banzhaf test $p$-value.

    Model-level scores are obtained by summing the per-variation scores across the five standard-anchor and five different-anchor variations:
    \[
    \mathrm{ABSS}_{m}=\sum_{v=1}^{10}\mathrm{ABSS}_{m,v}.
    \]
    The positive-control variation V0 is excluded from this aggregation. Because ABSS combines effect magnitude with significance-dependent weights, it is interpreted as a descriptive sensitivity index conditional on the present experimental design, rather than a population-level measure of model susceptibility.

\end{enumerate}

% =========================
\section{Results}
\label{sec:results}

\begin{figure}[!htbp]
    \centering
    \includegraphics[width=\textwidth]{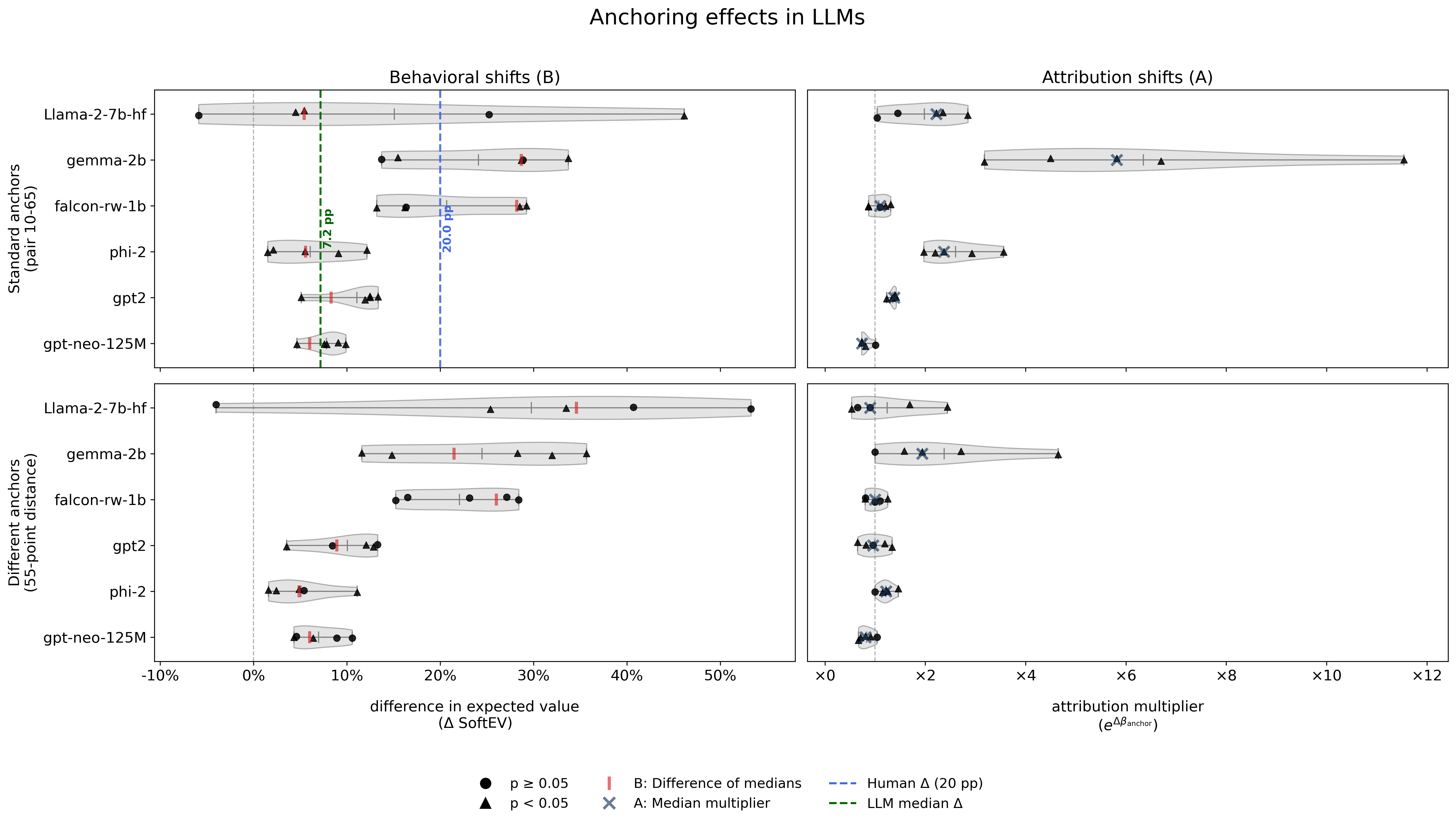}
    \caption{Behavioral and attributional anchoring effects across six LLMs. Grey violins summarize variation-level behavioral shifts ($\Delta\mathrm{SoftEV}$, in percentage points) and Banzhaf attribution multipliers ($e^{\Delta\beta_{\mathrm{anchor}}}$) across five non-control prompt variations per model and regime. Black circles and triangles indicate $p\geq0.05$ and $p<0.05$, respectively, from the corresponding target-level tests. Red vertical markers indicate model-specific differences between median high- and low-anchor SoftEV values, whereas blue-grey crosses indicate median attribution multipliers. In the standard-anchor behavioral panel, the blue dashed line marks the 20-percentage-point human anchoring effect reported by Tversky and Kahneman~\cite{doi:10.1126/science.185.4157.1124}, based on median estimates of 25\% and 45\% under anchors of 10 and 65, respectively. The green dashed line marks the median across the six model-specific differences of SoftEV medians (7.17 percentage points), giving each model equal weight. This comparison is descriptive: the human benchmark reflects participant responses to a single question, whereas the LLM statistic summarizes model-derived expected values across different questions. Positive-control variation (V0) excluded.}
    \label{fig:anchoring_summary}
\end{figure}

Figure~\ref{fig:anchoring_summary} summarizes behavioral (B) and attributional (A) anchoring effects across six LLMs under standard and different anchor regimes. Detailed per-variation results, including effect sizes, statistical significance, and Wilcoxon and permutation robustness checks, are reported in Table~\ref{tab:all_llms_by_model} (Appendix~\ref{app:sentinel}).

Behavioral shifts were predominantly positive across models, indicating that higher anchors generally increased expected estimates. However, their magnitudes varied substantially. \textit{gemma-2b} and \textit{falcon-rw-1b} exhibited relatively large positive shifts, whereas \textit{Llama-2-7b-hf} showed the greatest dispersion, including occasional reversals. \textit{gpt2}, \textit{gpt-neo-125M}, and \textit{phi-2} generally displayed smaller but predominantly positive effects. Target-level tests frequently yielded significant log-probability differences, providing conditional evidence of distributional reweighting over the evaluated targets.

Banzhaf attribution revealed greater directional heterogeneity. \textit{gemma-2b} produced the largest positive attribution multipliers, while \textit{phi-2} showed predominantly positive attributional shifts, including variations with comparatively small behavioral effects. \textit{Llama-2-7b-hf} exhibited several strong positive attributional responses alongside occasional reversals. \textit{gpt2} displayed mostly positive but smaller attributional effects, whereas \textit{falcon-rw-1b} showed mixed directions. Notably, \textit{gpt-neo-125M} frequently combined positive behavioral shifts with negative attributional changes. Thus, increases in expected estimates did not necessarily coincide with larger anchor-field Banzhaf contributions to the evaluated log-probabilities.

The standard-anchor regime additionally allows a descriptive comparison with the original human experiment. The median across model-specific differences of SoftEV medians was 7.17 percentage points, below the historical human benchmark of 20 percentage points reported by Tversky and Kahneman~\cite{doi:10.1126/science.185.4157.1124}. Nevertheless, individual models and prompt variations exhibited substantially larger shifts. Because the human benchmark reflects participant responses to one question, whereas the LLM statistic aggregates model-derived expectations across multiple questions, this comparison provides historical context rather than a direct measure of relative human--LLM anchoring susceptibility.

\subsection{Standard vs.\ different anchors}
\label{sec:standard_vs_different}

Changing the absolute anchor values while preserving their 55-point separation generally maintained positive behavioral shifts, although attributional alignment was less consistent (Figure~\ref{fig:anchoring_summary}). \textit{gemma-2b} and \textit{phi-2} exhibited broadly aligned positive behavioral and attributional responses across both regimes. \textit{gpt2} retained positive behavioral shifts, but some different-anchor variations produced negative Banzhaf attributional shifts. \textit{Llama-2-7b-hf} maintained large behavioral responses under different anchors while showing greater variability in attributional direction. \textit{falcon-rw-1b} exhibited positive behavioral shifts alongside mixed attributional responses in both regimes. Finally, \textit{gpt-neo-125M} remained predominantly discordant, with positive behavioral effects frequently accompanied by negative Banzhaf attributional shifts.

% \FloatBarrier

Overall, the direction of behavioral anchoring largely persisted when the numerical anchors changed, whereas attributional alignment varied more strongly across models and prompt variations. Similar directional behavioral responses therefore did not necessarily correspond to similar patterns of anchor-field attribution. However, the limited number of evaluated questions restricts the generalizability of these patterns to broader prompting contexts.

\subsection{Ranking: Anchoring Bias Sensitivity Score (ABSS)}
\label{sec:leaderboard_abss}

\begin{figure}[!htbp]
    \centering
    \includegraphics[width=\textwidth]{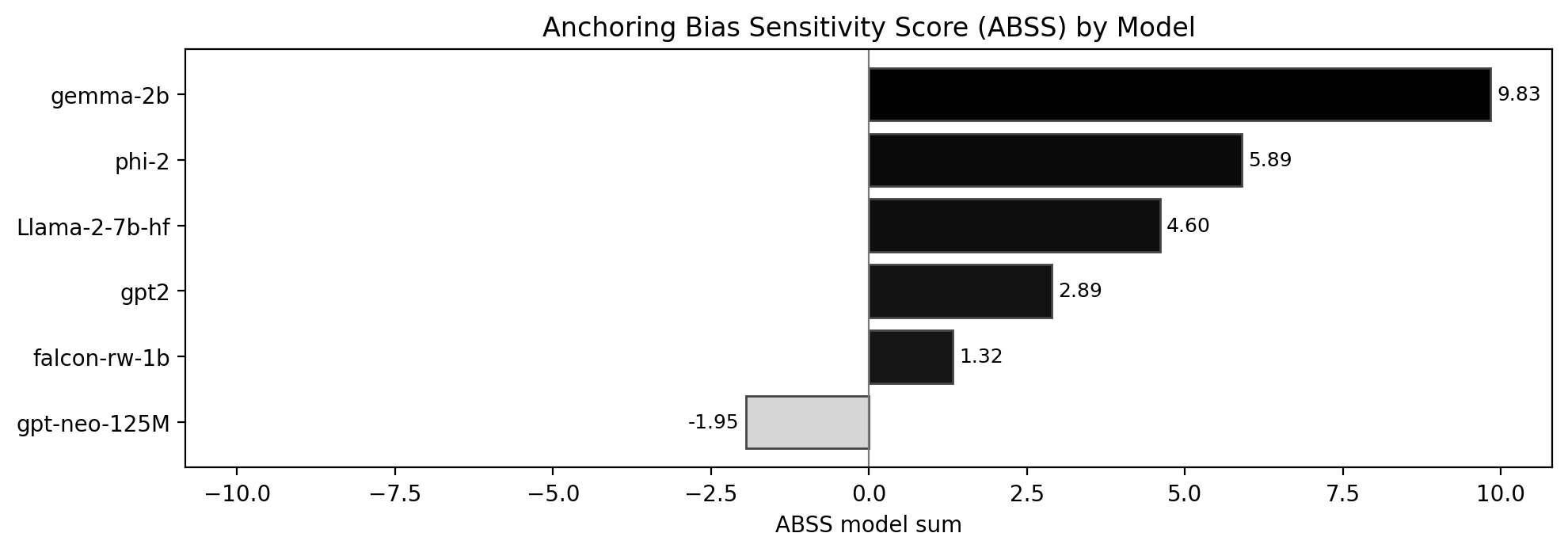}
    \caption{Anchoring Bias Sensitivity Score (ABSS) summed across ten non-control prompt variations per model. The composite combines directional behavioral effects, Banzhaf attributional shifts, significance-based weights, behavioral robustness diagnostics, and directional concordance. Higher positive scores indicate stronger aggregate alignment with the defined anchoring direction, whereas negative scores indicate that negative contributions dominate the composite. The positive-control variation (V0) is excluded.}
    \label{fig:abss_by_model}
\end{figure}

The ABSS combines behavioral and Banzhaf attributional evidence into a model-level composite, accounting for effect magnitude and direction, significance-based weights, behavioral robustness, and concordance. As shown in Figure~\ref{fig:abss_by_model}, \textit{gemma-2b} ranked first (9.83), followed by \textit{phi-2} (5.89) and \textit{Llama-2-7b-hf} (4.60). These models frequently combined positive behavioral shifts with positive Banzhaf attributional contributions.

\textit{gpt2} (2.89) and \textit{falcon-rw-1b} (1.32) ranked lower, reflecting less consistent joint behavioral--attributional alignment. \textit{gpt-neo-125M} received a negative aggregate score ($-1.95$) despite frequently exhibiting positive behavioral shifts, consistent with its predominantly negative attributional contributions.

% \FloatBarrier

These rankings demonstrate that behavioral anchoring magnitude alone does not determine ABSS. The composite also incorporates the direction and statistical evidence of attributional changes, behavioral robustness, and concordance between the two signals. Accordingly, the ranking provides a descriptive comparison of sensitivity within the evaluated experimental conditions, rather than a general ordering of model susceptibility to anchoring. Per-variation ABSS scores are presented in Figure~\ref{fig:abss_per_variation} (Appendix~\ref{app:abss}).

% =========================

\section{Limitations}
\label{sec:limitations}

The present design has six main limitations. First, \emph{output discretization}: model responses were evaluated over a fixed set of numerical strings, which may not capture the variability of free-form generations or generalize to other response formats.

Second, \emph{limited prompt coverage}: the analysis includes five non-control question variations per anchor regime and model. Although these variations probe sensitivity across different question contexts and anchor values, they do not represent the broader space of possible prompts. Consequently, model-level summaries and rankings may depend on the selected stimuli. The violin plots provide descriptive visualizations of these small sets of observations, rather than precise estimates of underlying population distributions.

Third, \emph{statistical dependence}: the behavioral and attributional tests operate on target-level log-probabilities and Banzhaf contributions derived from the same model evaluations. These quantities are not independent experimental observations, and conventional test assumptions may therefore be imperfectly satisfied. Accordingly, the reported significance and robustness tests should be interpreted as conditional diagnostics of differences across the evaluated targets, rather than independent-sample evidence generalizable to unseen prompts. Because ABSS incorporates significance-based weights, this limitation also affects the interpretation of its aggregate rankings.

Fourth, \emph{human comparability}: the historical human anchoring benchmark summarizes participant responses to a single question, whereas the aggregate LLM comparison is derived from model-implied expected values across multiple prompt variations. Although both use differences between condition-specific medians, their underlying observational units differ, limiting direct quantitative comparisons.

Fifth, \emph{attributional scope}: Banzhaf values quantify average marginal field contributions under the specified log-probability payoff and coalition procedure. Unlike classical Shapley values, they do not generally satisfy the efficiency property. Moreover, the analysis does not establish mechanistic explanations, as internal representations, parameters, and circuits were not investigated. The attributional results therefore characterize field-level influence under the chosen evaluation procedure rather than the internal mechanisms responsible for anchoring.

Finally, \emph{model accessibility}: the approach requires access to output log-probabilities and repeated evaluation of modified prompts, restricting its application to models and interfaces that expose these capabilities.

% =========================
\section{Discussion and Conclusion}
\label{sec:discussion}

The findings provide evidence that LLMs exhibit anchoring-like behavior, with higher numerical anchors generally shifting the probability distribution over candidate answers toward larger values (\textbf{B+}). However, the magnitude and consistency of these effects vary across models and prompt conditions. \textit{gemma-2b}, \textit{phi-2}, and \textit{Llama-2-7b-hf} achieve the highest Anchoring Bias Sensitivity Scores (ABSS), reflecting frequent positive behavioral shifts accompanied by aligned attributional contributions. In contrast, \textit{gpt2}, \textit{falcon-rw-1b}, and \textit{gpt-neo-125M} rank lower, although this does not necessarily imply weaker behavioral anchoring. For example, \textit{falcon-rw-1b} exhibits substantial behavioral shifts despite mixed attributional alignment, while \textit{gpt-neo-125M} frequently combines positive behavioral effects with negative attributional changes. These differences demonstrate that behavioral anchoring magnitude and attributional alignment capture distinct aspects of model sensitivity.

Methodologically, this study combines distribution-level behavioral analysis with exact Banzhaf attribution over structured prompt fields, using sequence log-probability as the payoff. Unlike classical Shapley values, Banzhaf values assign equal weight to the marginal contributions of a feature across coalitions. With only four prompt fields, all 16 coalitions can be evaluated exhaustively, producing deterministic attributions without Monte Carlo approximation. Integrating these contributions with behavioral shifts through ABSS provides a structured summary of anchoring sensitivity within the evaluated conditions. Importantly, attributional evidence identifies how the anchor field contributes to model log-probabilities under the specified payoff and coalition procedure, but does not establish the internal mechanisms responsible for these effects.

The findings have implications for both research streams introduced earlier. From the \emph{LLMs-as-subjects} perspective, including their use as synthetic participants or users, models reproduce anchoring-like patterns in controlled experiments, yet variation across prompts and differences in attributional alignment caution against treating them as direct substitutes for human participants. The historical comparison with Tversky and Kahneman~\cite{doi:10.1126/science.185.4157.1124} provides useful context, although the human and LLM statistics are not directly equivalent. From the \emph{LLMs-as-decision-systems} perspective, including applications such as \emph{LLM-as-a-judge} and automated evaluation, sensitivity to arbitrary numerical cues raises concerns about the reliability of model-assisted judgments, particularly in consequential domains such as healthcare, finance, and law. While the present experiments do not evaluate downstream harms, they illustrate a form of context sensitivity that warrants attention in model evaluation and deployment.

Several directions for future research follow. Expanding the number and diversity of prompts and models, including instruction-tuned and frontier systems with log-probability access, would help assess the generalizability of these findings. Integrating Banzhaf attribution with concept-based or mechanistic interpretability could further investigate the relationship between observable probability shifts, input-field contributions, and internal representations. The framework could also be extended to other cognitive biases, such as framing and availability, and used to evaluate potential mitigation strategies under more realistic decision-making conditions. Such extensions would help establish whether behavioral--attributional relationships persist across tasks, model families, and deployment settings.

Overall, this study demonstrates that numerical anchors can systematically influence LLM output distributions and that these behavioral shifts are often, but not always, accompanied by aligned Banzhaf attributional changes. By combining log-probability-based behavioral measurement with exact field-level attribution and an integrated sensitivity score, the proposed framework provides a reproducible approach to examining anchoring in models with accessible log-probabilities. The results highlight the importance of evaluating both observable behavior and input contributions when investigating cognitive-bias-like effects, while providing a foundation for further research connecting behavioral science, explainable AI, and AI safety.

% =========================
\section{Data Availability}
\label{sec:reproducibility}

To facilitate transparency and replication, the GitHub repository at \href{https://github.com/felipevalencla/Anchoring-LLMs}{\texttt{Anchoring-LLMs}} contains the code, processed datasets, prompt templates, and supplementary documentation.

% =========================
\bibliographystyle{unsrt}
\bibliography{references} % references.bib

% =========================

% =========================
\appendix
\section{Prompt Templates \& Anchor Lists}
\label{app:prompts}

\subsection*{Standard-anchors set (pair 10--65; different questions, same structure)}

\begin{itemize}[leftmargin=*]
    \item \textbf{S1.} \emph{The spinner stopped at} \(\;\vert\) \emph{Do you think the percentage of Asian countries in the UN is higher or lower than} \(\;\vert\) \emph{Guess the percentage of Asian countries in the UN.} \quad Anchors: 10, 65.
    \item \textbf{S2.} \emph{You saw a wheel display} \(\;\vert\) \emph{Would you say the percentage of South American countries in the UN is more or less than} \(\;\vert\) \emph{Estimate the percentage of South American countries in the UN.} \quad Anchors: 10, 65.
    \item \textbf{S3.} \emph{The number you received was} \(\;\vert\) \emph{Is the share of English-speaking countries in the UN larger or smaller than} \(\;\vert\) \emph{Provide your best estimate of their percentage.} \quad Anchors: 10, 65.
    \item \textbf{S4.} \emph{The game spun a number:} \(\;\vert\) \emph{Is the proportion of EU member countries in the UN greater or lesser than} \(\;\vert\) \emph{What do you think is their percentage in the UN?} \quad Anchors: 10, 65.
    \item \textbf{S5.} \emph{Imagine you just saw the number} \(\;\vert\) \emph{Would you say the percentage of French-speaking countries in the UN is above or below} \(\;\vert\) \emph{Estimate the percentage of French-speaking countries in the UN.} \quad Anchors: 10, 65.
\end{itemize}

\subsection*{Different-anchors set (constant 55-point distance; different questions, same structure)}

The different-anchors set uses the same question-specific prompt fields as the corresponding standard-anchors variations, replacing only the numerical anchor values.

\begin{itemize}[leftmargin=*]
    \item \textbf{D1.} Asian countries in the UN (same prompt fields as S1). \quad Anchors: 15, 70.
    \item \textbf{D2.} South American countries in the UN (same prompt fields as S2). \quad Anchors: 20, 75.
    \item \textbf{D3.} English-speaking countries in the UN (same prompt fields as S3). \quad Anchors: 25, 80.
    \item \textbf{D4.} EU member countries in the UN (same prompt fields as S4). \quad Anchors: 30, 85.
    \item \textbf{D5.} French-speaking countries in the UN (same prompt fields as S5). \quad Anchors: 35, 90.
\end{itemize}

\paragraph{Design note.}
All prompt variations follow the four-field template described in Methods. The vertical bars above separate the scene, comparative, and absolute-question components; the numerical anchor is inserted into the scene and comparative fields during rendering. Maintaining a constant 55-point difference between low and high anchors facilitates comparisons across anchor pairs, while varying their absolute values reduces reliance on a single potentially familiar numerical pair. However, these manipulations do not eliminate potential training-data contamination. The positive-control variation (V0) is retained for reference but excluded from aggregate analyses and the Anchoring Bias Sensitivity Score (ABSS).

% =========================
\section{Anchoring Results Grouped by Model}
\label{app:sentinel}

\label{sec:all_llms_overview}

% Include the longtable first so the label is defined:
% --- Unicode mappings for pdfLaTeX (keep if needed) ---
\providecommand{\DeclareUnicodeCharacter}[2]{}

% --- Column types (requires \usepackage{array} in the preamble) ---
\newcolumntype{L}[1]{>{\raggedright\arraybackslash}p{#1}}
\newcolumntype{C}[1]{>{\centering\arraybackslash}p{#1}}
\newcolumntype{R}[1]{>{\raggedleft\arraybackslash}p{#1}}

% (optional, helps long captions) \setlength\LTcapwidth{\textwidth}
% (optional, let body use full width) \setlength\LTleft{0pt}\setlength\LTright{0pt}

\small % switch to \footnotesize if still too wide
\begin{longtable}{
  L{1.2cm}  % Model
  L{0.8cm}  % Var
  L{0.9cm}  % Anchors
  C{1.6cm}  % SoftEV (Low)
  C{1.6cm}  % SoftEV (High)
  R{1.1cm}  % ΔEV
  L{1.0cm}  % B (EV dir; t-test on logprobs)
  L{0.8cm}  % R (Wilcoxon / Perm)
  L{0.9cm}  % A (ΔBanzhaf; t-test)
  C{1.9cm}  % ΔBanzhaf [nats] (×mult)
}

\caption{Anchoring results grouped by model, showing all variations per LLM.\\
SoftEV: point estimate and 95\% predictive interval (bootstrap).\\
$\Delta\mathrm{EV}$: high-minus-low SoftEV difference.\\
B: behavioral direction ($\Delta\mathrm{EV}$) with paired t-test on target-level log-probabilities.\\
R: behavioral robustness diagnostics (Wilcoxon = W, permutation = P).\\
A: attributional direction ($\Delta\beta$) with paired t-test on target-level Banzhaf values.\\
$\Delta\beta$: high-minus-low mean anchor-field Banzhaf attribution (nats); multiplier $= e^{\Delta\beta}$.}
\label{tab:all_llms_by_model} \\

\toprule
\textbf{Model} &
\textbf{Var} &
\textbf{Anchors} &
\begin{tabular}{@{}c@{}}\textbf{SoftEV}\\\textbf{(Low)}\\\scriptsize[95\% pred]\end{tabular} &
\begin{tabular}{@{}c@{}}\textbf{SoftEV}\\\textbf{(High)}\\\scriptsize[95\% pred]\end{tabular} &
\boldmath$\Delta$\unboldmath\textbf{EV} &
\begin{tabular}{@{}c@{}}\textbf{B}\end{tabular} &
\begin{tabular}{@{}c@{}}\textbf{R}\end{tabular} &
\begin{tabular}{@{}c@{}}\textbf{A}\end{tabular} &
\begin{tabular}{@{}c@{}}\textbf{$\boldsymbol{\Delta\beta}$}\\\scriptsize[nats] ($\times$mult)\end{tabular} \\
\midrule
\endfirsthead

\toprule
\textbf{Model} &
\textbf{Var} &
\textbf{Anchors} &
\begin{tabular}{@{}c@{}}\textbf{SoftEV}\\\textbf{(Low)}\\\scriptsize[95\% pred]\end{tabular} &
\begin{tabular}{@{}c@{}}\textbf{SoftEV}\\\textbf{(High)}\\\scriptsize[95\% pred]\end{tabular} &
\boldmath$\Delta$\unboldmath\textbf{EV} &
\begin{tabular}{@{}c@{}}\textbf{B}\end{tabular} &
\begin{tabular}{@{}c@{}}\textbf{R}\end{tabular} &
\begin{tabular}{@{}c@{}}\textbf{A}\end{tabular} &
\begin{tabular}{@{}c@{}}\textbf{$\boldsymbol{\Delta\beta}$}\\\scriptsize[nats] ($\times$mult)\end{tabular} \\
\midrule
\endhead

\midrule
\multicolumn{10}{r}{\footnotesize Continued on next page} \\
\midrule
\endfoot

\bottomrule
\endlastfoot

% ---- Your 66 data rows go here (unchanged). Do not use \hline anywhere. ----
\multirow{11}{*}\scriptsize{Llama-2-7b-hf} & V0 & 10→65 & 0.11 \scriptsize[0.00-0.90] & 62.95 \scriptsize[61.24-64.35] & +62.84 & \scriptsize{B+***} & \scriptsize{W*** P***} & \scriptsize{A+***} & 0.50 (×1.65) \\
 & V1-S & 10→65 & 39.61 \scriptsize[31.40-47.43] & 64.85 \scriptsize[63.44-65.87] & +25.24 & \scriptsize{B+} & \scriptsize{W** P} & \scriptsize{A+} & 0.04 (×1.04) \\
 & V1-D & 15→70 & 77.86 \scriptsize[72.39-82.75] & 73.84 \scriptsize[71.17-76.27] & -4.01 & \scriptsize{B–} & \scriptsize{W* P} & \scriptsize{A–***} & -0.63 (×0.53) \\
 & V2-S & 10→65 & 61.30 \scriptsize[54.28-67.94] & 65.81 \scriptsize[63.49-68.02] & +4.51 & \scriptsize{B+**} & \scriptsize{W*** P**} & \scriptsize{A+*} & 0.37 (×1.45) \\
 & V2-D & 20→75 & 35.05 \scriptsize[29.49-40.98] & 75.74 \scriptsize[73.91-77.34] & +40.69 & \scriptsize{B+} & \scriptsize{W P} & \scriptsize{A–} & -0.11 (×0.90) \\
 & V3-S & 10→65 & 67.43 \scriptsize[61.77-72.59] & 61.53 \scriptsize[58.49-64.22] & -5.89 & \scriptsize{B–} & \scriptsize{W P} & \scriptsize{A+***} & 0.80 (×2.22) \\
 & V3-D & 25→80 & 41.15 \scriptsize[35.70-46.90] & 74.65 \scriptsize[70.98-77.62] & +33.49 & \scriptsize{B+***} & \scriptsize{W*** P***} & \scriptsize{A+***} & 0.89 (×2.44) \\
 & V4-S & 10→65 & 18.19 \scriptsize[14.15-22.57] & 64.33 \scriptsize[63.25-65.01] & +46.15 & \scriptsize{B+***} & \scriptsize{W*** P***} & \scriptsize{A+***} & 0.85 (×2.35) \\
 & V4-D & 30→85 & 30.88 \scriptsize[30.05-31.80] & 84.16 \scriptsize[82.83-85.04] & +53.29 & \scriptsize{B+} & \scriptsize{W P} & \scriptsize{A–} & -0.42 (×0.65) \\
 & V5-S & 10→65 & 59.25 \scriptsize[51.96-66.11] & 64.68 \scriptsize[62.71-66.36] & +5.43 & \scriptsize{B+***} & \scriptsize{W*** P***} & \scriptsize{A+***} & 1.05 (×2.85) \\
 & V5-D & 35→90 & 61.76 \scriptsize[56.17-67.09] & 87.14 \scriptsize[84.58-89.18] & +25.38 & \scriptsize{B+***} & \scriptsize{W*** P***} & \scriptsize{A+***} & 0.52 (×1.69) \\
\multirow{11}{*}\scriptsize{falcon-rw-1b} & V0 & 10→65 & 37.97 \scriptsize[31.40-44.54] & 49.84 \scriptsize[43.95-55.43] & +11.86 & \scriptsize{B+} & \scriptsize{W P} & \scriptsize{A+} & 0.03 (×1.03) \\
 & V1-S & 10→65 & 43.34 \scriptsize[36.74-49.80] & 59.57 \scriptsize[54.67-63.94] & +16.23 & \scriptsize{B+***} & \scriptsize{W*** P***} & \scriptsize{A+} & 0.09 (×1.10) \\
 & V1-D & 15→70 & 37.89 \scriptsize[31.97-43.72] & 65.00 \scriptsize[60.64-68.82] & +27.11 & \scriptsize{B+*} & \scriptsize{W** P*} & \scriptsize{A–***} & -0.22 (×0.80) \\
 & V2-S & 10→65 & 45.57 \scriptsize[38.79-52.09] & 61.90 \scriptsize[56.82-66.53] & +16.33 & \scriptsize{B+} & \scriptsize{W P} & \scriptsize{A–**} & -0.14 (×0.87) \\
 & V2-D & 20→75 & 47.38 \scriptsize[41.00-53.45] & 63.90 \scriptsize[58.47-68.86] & +16.51 & \scriptsize{B+} & \scriptsize{W P} & \scriptsize{A+} & 0.00 (×1.00) \\
 & V3-S & 10→65 & 24.76 \scriptsize[19.70-30.34] & 53.99 \scriptsize[48.45-59.21] & +29.24 & \scriptsize{B+***} & \scriptsize{W*** P***} & \scriptsize{A+***} & 0.27 (×1.31) \\
 & V3-D & 25→80 & 30.54 \scriptsize[26.45-34.88] & 53.67 \scriptsize[47.16-59.80] & +23.13 & \scriptsize{B+} & \scriptsize{W P} & \scriptsize{A+**} & 0.22 (×1.25) \\
 & V4-S & 10→65 & 30.26 \scriptsize[24.07-36.66] & 43.45 \scriptsize[37.18-49.65] & +13.19 & \scriptsize{B+***} & \scriptsize{W*** P***} & \scriptsize{A–**} & -0.14 (×0.87) \\
 & V4-D & 30→85 & 31.87 \scriptsize[26.42-37.66] & 47.09 \scriptsize[39.70-54.17] & +15.22 & \scriptsize{B+} & \scriptsize{W** P} & \scriptsize{A+} & 0.10 (×1.10) \\
 & V5-S & 10→65 & 31.38 \scriptsize[25.31-37.64] & 59.92 \scriptsize[54.64-64.66] & +28.53 & \scriptsize{B+***} & \scriptsize{W*** P***} & \scriptsize{A+**} & 0.19 (×1.21) \\
 & V5-D & 35→90 & 41.03 \scriptsize[36.56-45.62] & 69.44 \scriptsize[63.21-75.08] & +28.41 & \scriptsize{B+} & \scriptsize{W P} & \scriptsize{A–*} & -0.21 (×0.81) \\
\multirow{11}{*}\scriptsize{gemma-2b} & V0 & 10→65 & 25.70 \scriptsize[20.68-31.04] & 52.45 \scriptsize[47.30-57.25] & +26.75 & \scriptsize{B+***} & \scriptsize{W*** P***} & \scriptsize{A+***} & 1.78 (×5.91) \\
 & V1-S & 10→65 & 35.97 \scriptsize[30.19-41.85] & 51.44 \scriptsize[45.61-57.05] & +15.47 & \scriptsize{B+***} & \scriptsize{W*** P***} & \scriptsize{A+***} & 1.16 (×3.18) \\
 & V1-D & 15→70 & 38.74 \scriptsize[32.78-44.63] & 53.56 \scriptsize[47.50-59.35] & +14.82 & \scriptsize{B+***} & \scriptsize{W*** P***} & \scriptsize{A+***} & 0.46 (×1.58) \\
 & V2-S & 10→65 & 23.43 \scriptsize[18.24-29.16] & 52.13 \scriptsize[46.48-57.46] & +28.70 & \scriptsize{B+***} & \scriptsize{W*** P***} & \scriptsize{A+***} & 2.45 (×11.54) \\
 & V2-D & 20→75 & 28.85 \scriptsize[23.98-34.07] & 57.13 \scriptsize[50.89-62.88] & +28.28 & \scriptsize{B+***} & \scriptsize{W*** P***} & \scriptsize{A+***} & 1.00 (×2.71) \\
 & V3-S & 10→65 & 39.07 \scriptsize[32.89-45.17] & 52.78 \scriptsize[47.16-58.16] & +13.71 & \scriptsize{B+} & \scriptsize{W P} & \scriptsize{A+***} & 1.50 (×4.50) \\
 & V3-D & 25→80 & 42.33 \scriptsize[36.51-47.99] & 53.91 \scriptsize[47.63-59.92] & +11.58 & \scriptsize{B+***} & \scriptsize{W*** P***} & \scriptsize{A+***} & 0.67 (×1.94) \\
 & V4-S & 10→65 & 22.12 \scriptsize[17.33-27.48] & 51.00 \scriptsize[46.34-55.27] & +28.88 & \scriptsize{B+*} & \scriptsize{W* P*} & \scriptsize{A+***} & 1.76 (×5.82) \\
 & V4-D & 30→85 & 31.28 \scriptsize[27.68-35.06] & 63.27 \scriptsize[57.21-68.79] & +31.99 & \scriptsize{B+***} & \scriptsize{W*** P***} & \scriptsize{A+***} & 1.54 (×4.65) \\
 & V5-S & 10→65 & 21.31 \scriptsize[16.42-26.66] & 55.03 \scriptsize[50.13-59.48] & +33.72 & \scriptsize{B+***} & \scriptsize{W*** P***} & \scriptsize{A+***} & 1.90 (×6.70) \\
 & V5-D & 35→90 & 35.66 \scriptsize[31.84-39.61] & 71.34 \scriptsize[64.73-77.19] & +35.68 & \scriptsize{B+***} & \scriptsize{W*** P***} & \scriptsize{A–} & -0.00 (×1.00) \\
\multirow{11}{*}\scriptsize{gpt-neo-125M} & V0 & 10→65 & 32.05 \scriptsize[25.99-38.29] & 42.50 \scriptsize[36.02-48.97] & +10.45 & \scriptsize{B+***} & \scriptsize{W*** P***} & \scriptsize{A–***} & -0.35 (×0.70) \\
 & V1-S & 10→65 & 46.86 \scriptsize[39.72-53.86] & 55.93 \scriptsize[49.52-62.08] & +9.07 & \scriptsize{B+***} & \scriptsize{W*** P***} & \scriptsize{A–***} & -0.32 (×0.73) \\
 & V1-D & 15→70 & 44.30 \scriptsize[37.66-50.70] & 54.87 \scriptsize[48.09-61.41] & +10.57 & \scriptsize{B+} & \scriptsize{W P} & \scriptsize{A–***} & -0.21 (×0.81) \\
 & V2-S & 10→65 & 48.24 \scriptsize[40.67-55.55] & 52.88 \scriptsize[45.63-59.92] & +4.64 & \scriptsize{B+***} & \scriptsize{W*** P***} & \scriptsize{A–***} & -0.31 (×0.73) \\
 & V2-D & 20→75 & 47.36 \scriptsize[40.04-54.41] & 51.94 \scriptsize[44.61-59.08] & +4.58 & \scriptsize{B+} & \scriptsize{W P} & \scriptsize{A–***} & -0.35 (×0.71) \\
 & V3-S & 10→65 & 44.25 \scriptsize[36.94-51.36] & 51.84 \scriptsize[44.93-58.48] & +7.59 & \scriptsize{B+***} & \scriptsize{W*** P***} & \scriptsize{A–***} & -0.22 (×0.81) \\
 & V3-D & 25→80 & 45.92 \scriptsize[39.04-52.69] & 50.24 \scriptsize[43.16-57.10] & +4.32 & \scriptsize{B+***} & \scriptsize{W*** P***} & \scriptsize{A+} & 0.04 (×1.04) \\
 & V4-S & 10→65 & 39.74 \scriptsize[33.89-45.53] & 49.64 \scriptsize[44.05-55.03] & +9.89 & \scriptsize{B+***} & \scriptsize{W*** P***} & \scriptsize{A–***} & -0.30 (×0.74) \\
 & V4-D & 30→85 & 41.49 \scriptsize[36.14-46.74] & 50.40 \scriptsize[44.40-56.19] & +8.91 & \scriptsize{B+} & \scriptsize{W P} & \scriptsize{A–***} & -0.40 (×0.67) \\
 & V5-S & 10→65 & 52.37 \scriptsize[44.74-59.67] & 60.21 \scriptsize[53.24-66.78] & +7.84 & \scriptsize{B+***} & \scriptsize{W*** P***} & \scriptsize{A+} & 0.01 (×1.01) \\
 & V5-D & 35→90 & 55.03 \scriptsize[47.90-61.87] & 61.43 \scriptsize[53.96-68.45] & +6.40 & \scriptsize{B+**} & \scriptsize{W* P*} & \scriptsize{A–**} & -0.09 (×0.91) \\
\multirow{11}{*}\scriptsize{gpt2} & V0 & 10→65 & 39.73 \scriptsize[32.56-46.75] & 44.60 \scriptsize[37.49-51.49] & +4.87 & \scriptsize{B+} & \scriptsize{W P} & \scriptsize{A+***} & 0.24 (×1.27) \\
 & V1-S & 10→65 & 40.18 \scriptsize[33.86-46.37] & 52.67 \scriptsize[47.07-57.91] & +12.49 & \scriptsize{B+***} & \scriptsize{W*** P***} & \scriptsize{A+***} & 0.29 (×1.34) \\
 & V1-D & 15→70 & 39.52 \scriptsize[33.76-45.17] & 52.36 \scriptsize[46.50-57.96] & +12.84 & \scriptsize{B+**} & \scriptsize{W* P**} & \scriptsize{A–} & -0.04 (×0.96) \\
 & V2-S & 10→65 & 45.01 \scriptsize[38.42-51.43] & 57.48 \scriptsize[51.75-62.94] & +12.47 & \scriptsize{B+***} & \scriptsize{W*** P***} & \scriptsize{A+***} & 0.21 (×1.23) \\
 & V2-D & 20→75 & 45.40 \scriptsize[39.41-51.33] & 58.69 \scriptsize[52.84-64.20] & +13.29 & \scriptsize{B+} & \scriptsize{W* P} & \scriptsize{A+***} & 0.17 (×1.19) \\
 & V3-S & 10→65 & 44.73 \scriptsize[38.00-51.32] & 56.67 \scriptsize[50.83-62.19] & +11.95 & \scriptsize{B+***} & \scriptsize{W*** P***} & \scriptsize{A+***} & 0.33 (×1.39) \\
 & V3-D & 25→80 & 47.23 \scriptsize[41.21-53.13] & 55.69 \scriptsize[49.34-61.69] & +8.46 & \scriptsize{B+} & \scriptsize{W P} & \scriptsize{A–***} & -0.20 (×0.82) \\
 & V4-S & 10→65 & 47.93 \scriptsize[41.50-54.25] & 53.04 \scriptsize[46.80-59.05] & +5.11 & \scriptsize{B+***} & \scriptsize{W*** P***} & \scriptsize{A+***} & 0.35 (×1.42) \\
 & V4-D & 30→85 & 48.87 \scriptsize[42.51-55.09] & 52.43 \scriptsize[46.19-58.52] & +3.55 & \scriptsize{B+***} & \scriptsize{W*** P***} & \scriptsize{A+***} & 0.29 (×1.34) \\
 & V5-S & 10→65 & 37.50 \scriptsize[30.75-44.09] & 50.85 \scriptsize[44.51-56.95] & +13.35 & \scriptsize{B+***} & \scriptsize{W*** P***} & \scriptsize{A+***} & 0.32 (×1.38) \\
 & V5-D & 35→90 & 42.25 \scriptsize[36.23-48.15] & 54.32 \scriptsize[47.22-61.11] & +12.07 & \scriptsize{B+***} & \scriptsize{W*** P***} & \scriptsize{A–***} & -0.43 (×0.65) \\
\multirow{11}{*}\scriptsize{phi-2} & V0 & 10→65 & 24.45 \scriptsize[19.51-29.80] & 33.95 \scriptsize[28.25-39.68] & +9.49 & \scriptsize{B+*} & \scriptsize{W* P*} & \scriptsize{A+***} & 0.52 (×1.69) \\
 & V1-S & 10→65 & 41.76 \scriptsize[35.68-47.73] & 53.91 \scriptsize[48.73-58.72] & +12.14 & \scriptsize{B+***} & \scriptsize{W*** P***} & \scriptsize{A+***} & 0.68 (×1.97) \\
 & V1-D & 15→70 & 40.06 \scriptsize[34.49-45.64] & 51.16 \scriptsize[45.45-56.65] & +11.09 & \scriptsize{B+***} & \scriptsize{W*** P***} & \scriptsize{A+**} & 0.14 (×1.15) \\
 & V2-S & 10→65 & 31.48 \scriptsize[26.50-36.69] & 32.99 \scriptsize[27.77-38.27] & +1.51 & \scriptsize{B+***} & \scriptsize{W*** P***} & \scriptsize{A+***} & 1.07 (×2.93) \\
 & V2-D & 20→75 & 30.79 \scriptsize[26.14-35.62] & 32.39 \scriptsize[27.24-37.69] & +1.60 & \scriptsize{B+***} & \scriptsize{W*** P***} & \scriptsize{A–} & -0.00 (×1.00) \\
 & V3-S & 10→65 & 21.34 \scriptsize[17.18-25.89] & 23.46 \scriptsize[18.69-28.52] & +2.12 & \scriptsize{B+***} & \scriptsize{W*** P***} & \scriptsize{A+***} & 1.27 (×3.56) \\
 & V3-D & 25→80 & 22.60 \scriptsize[18.11-27.42] & 25.02 \scriptsize[20.08-30.23] & +2.42 & \scriptsize{B+***} & \scriptsize{W*** P***} & \scriptsize{A+***} & 0.20 (×1.22) \\
 & V4-S & 10→65 & 18.13 \scriptsize[14.39-22.21] & 27.22 \scriptsize[22.27-32.28] & +9.09 & \scriptsize{B+***} & \scriptsize{W*** P***} & \scriptsize{A+***} & 0.79 (×2.20) \\
 & V4-D & 30→85 & 23.42 \scriptsize[19.65-27.43] & 28.83 \scriptsize[23.66-34.20] & +5.41 & \scriptsize{B+} & \scriptsize{W P} & \scriptsize{A+***} & 0.38 (×1.46) \\
 & V5-S & 10→65 & 21.64 \scriptsize[17.60-25.92] & 27.16 \scriptsize[22.55-31.97] & +5.52 & \scriptsize{B+***} & \scriptsize{W*** P***} & \scriptsize{A+***} & 0.86 (×2.37) \\
 & V5-D & 35→90 & 24.68 \scriptsize[20.80-28.70] & 29.56 \scriptsize[24.79-34.57] & +4.88 & \scriptsize{B+**} & \scriptsize{W* P**} & \scriptsize{A+***} & 0.22 (×1.25) \\
\end{longtable}

% =========================
% Appendix: Illustrative 3D distribution space
% =========================
\clearpage
\section{Behavioral and Attributional Distribution Space}
\label{app:behavior_attribution_example}

\begin{figure}[h]
    \centering
    \includegraphics[width=0.95\textwidth,height=0.78\textheight,keepaspectratio]{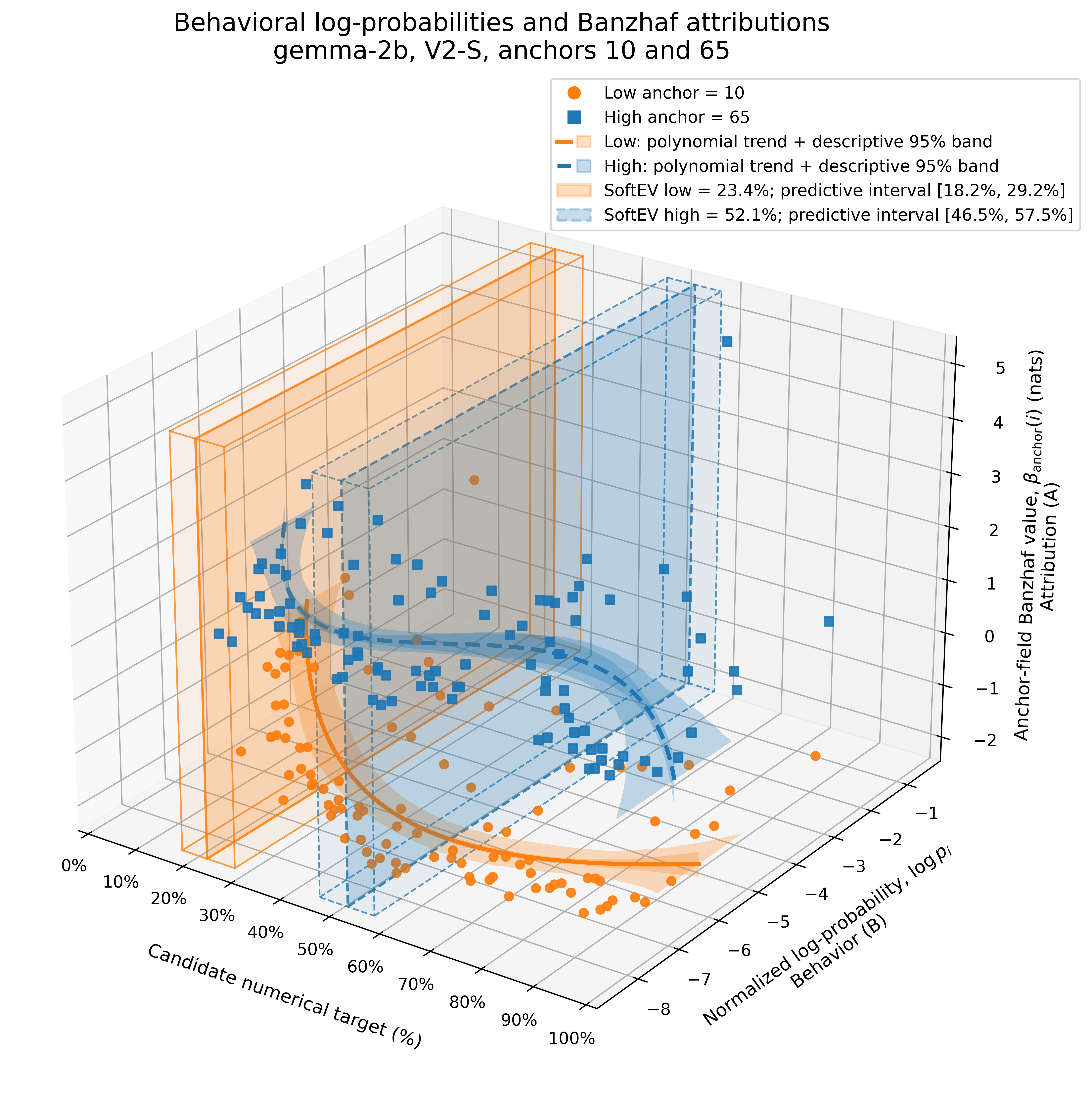}
    \caption{Three-dimensional visualization of normalized target-level log-probabilities (B) and anchor-field Banzhaf values (A) under low (10, orange circles) and high (65, blue squares) anchors. Polynomial curves and shaded ribbons represent descriptive fitted trends and bootstrap bands. Vertical planes indicate SoftEV estimates, while translucent slabs represent 95\% predictive intervals for the mean of 100 simulated draws.}
    \label{fig:behavior_attribution_3d}
\end{figure}

Figure~\ref{fig:behavior_attribution_3d} presents an illustrative example using \textit{gemma-2b} (V2-S), where the behavioral shift is $\Delta\mathrm{EV}=+28.70$ percentage points and the attributional shift is $\Delta\beta=+2.45$ nats (multiplier $\times 11.54$). This example is descriptive and does not establish a causal or mechanistic relationship between the two measurements.

% --- Appendix: ABSS per variation (landscape) ---
\section{ABSS per Variation}
\label{app:abss}

\begin{sidewaysfigure}[p]
    \centering
    \includegraphics[width=1\textheight,keepaspectratio]{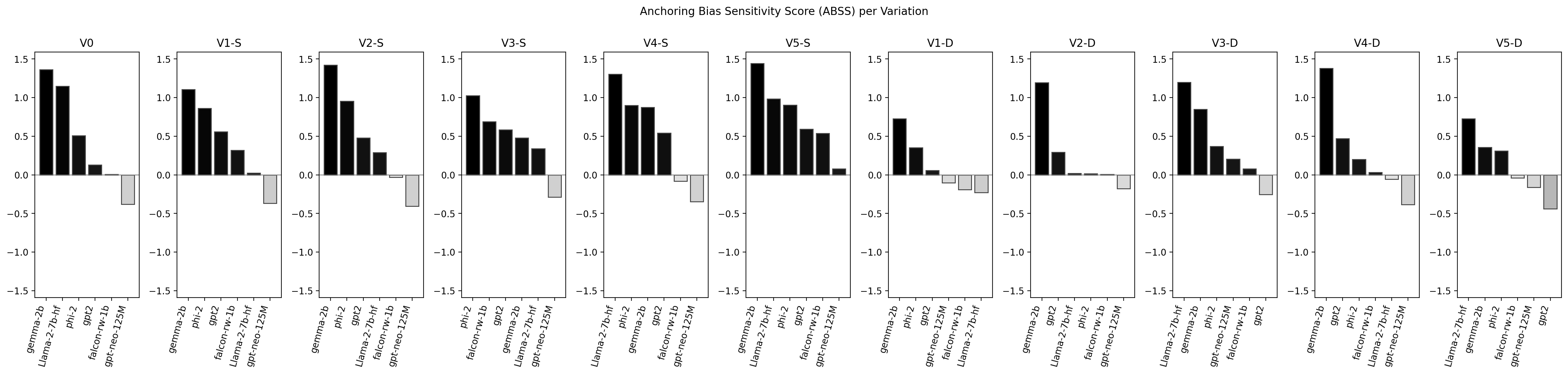}
    \caption{Anchoring Bias Sensitivity Score (ABSS) per variation across models.}
    \label{fig:abss_per_variation}
\end{sidewaysfigure}

% =========================

\end{document}